\documentclass[11pt]{article}

\usepackage[preprint]{acl}

\usepackage{times}
\usepackage{latexsym}
\usepackage[T1]{fontenc}
\usepackage[utf8]{inputenc}
\usepackage{microtype}
\usepackage{inconsolata}

\usepackage{booktabs}
\usepackage{array}
\usepackage{multirow}
\usepackage{graphicx}
\usepackage{tikz}
\usetikzlibrary{arrows.meta,positioning,fit,calc,decorations.pathreplacing}
\usepackage{amsmath}
\usepackage{url}

\title{ BharatSetu: A 1M$\times$22 Bridge for Indic Translation}

\author{Haq Nawaz Malik*\\
  {\small\texttt{ORCID:  0009-0003-1994-7640}}
  \And
  Nahfid Nissar\\
  {\small\texttt{ORCID:  0009-0002-2805-4687}}
  \And
  Faizan Iqbal\\
  {\small\texttt{ORCID: 0009-0002-8998-9347}}}

\begin{document}
\maketitle

\begin{abstract}
India's Eighth Schedule recognises 22 languages, yet parallel data is uneven across them, and many language pairs, especially among lower-resource languages such as Bodo, Dogri, Santali and Kashmiri, lack direct bitext. We present BharatSetu 22-Indic 1M, a row-aligned table of 984,764 English sentences, each paired with machine translations into all 22 scheduled languages (21,664,808 Indic cells), generated with the \texttt{sarvamai/sarvam-translate} model family. Because every Indic cell in a row translates the same English sentence, any two columns form a translation pair: 462 directed Indic-to-Indic pairs, or 506 including English, without direct bitext. We call these \emph{pivot-derived pairs} and propose a confidence layer that lets users choose how strictly each pair is connected. The corpus is machine-generated and not human-verified; quality is expected to vary by language, domain and sentence style, and pivot-derived pairs can inherit errors from both sides. We therefore release BharatSetu as a machine layer and propose a human-correction layer keyed to the same row identifiers, so that each corrected column upgrades every pair that uses it, together with an evaluation protocol. We report no translation-quality results.
\end{abstract}

\section{Introduction}
\label{sec:introduction}
This section states the resource gap that BharatSetu addresses and defines the contribution made by a single row-aligned table across all scheduled Indian languages. The Eighth Schedule of the Constitution of India lists 22 languages; the Government of India lists Assamese, Bengali, Bodo, Dogri, Gujarati, Hindi, Kannada, Kashmiri, Konkani, Maithili, Malayalam, Manipuri, Marathi, Nepali, Odia, Punjabi, Sanskrit, Santali, Sindhi, Tamil, Telugu and Urdu as the languages included in the Schedule \citep{govindia2026eighthschedule}. Modern machine translation (MT) resources, however, do not cover these languages with equal density. Widely used Indic corpora and models have improved coverage substantially, but many resources still concentrate on higher-resource languages or expose data as separate bilingual files rather than as one uniform multiway table \citep{kakwani2020indicnlpsuite,ramesh2022samanantar,gala2023indictrans2}.

The practical problem is not only that some individual languages have limited data. It is also that researchers often lack a common sentence spine across all 22 scheduled languages. Without such a spine, comparisons across languages require reconciling different domains, sentence selections, identifiers, preprocessing conventions and file formats. These differences make it difficult to train one model under controlled conditions, to study transfer from higher-resource to lower-resource languages, or to create pivot-derived pairs that share the same underlying meaning.

BharatSetu 22-Indic 1M addresses this problem by packaging 984,764 English source rows with one aligned machine-translated target column for each of the 22 scheduled languages. In scale terms, this is approximately 1M English rows $\times$ 22 Indic columns, producing 21,664,808 machine-generated Indic cells, or about 21.7M target-language cells. Each row has an \texttt{id} aligned to the original source corpus, the English source sentence, a \texttt{source} provenance label and 22 full-language-name target columns. The dataset is distributed publically open-source through the Hugging Face repository \href{https://huggingface.co/datasets/Omarrran/BharatSetu-22Indic-1M}{\texttt{Omarrran/BharatSetu-22Indic-1M}} \citep{bharatsetu2026}. 

The central claim of this paper is deliberately narrow. BharatSetu is valuable because it combines breadth, uniformity and scale in one table. It is not presented as a human-verified gold corpus, and we do not report BLEU, chrF, COMET or human-evaluation results. The translations were generated with the \texttt{sarvamai/sarvam-translate} model family, and the released corpus should be understood as a machine layer whose utility must be tested and improved through the evaluation and correction plan described in Sections~\ref{sec:evaluation} and~\ref{sec:limitations}.

\noindent Our contributions are:
\begin{itemize}
  \item \textbf{The need: one table for all 22 languages.} We release BharatSetu,
  984{,}764 English sentences aligned to machine translations in all 22 scheduled
  languages, including languages with little parallel data such as Bodo, Dogri,
  Santali and Kashmiri.
  \item \textbf{The use: any Indic language to any other.} Because every language
  translates the same English sentence, any two columns form a translation pair:
  462 Indic-to-Indic directions with no direct bitext, each with a confidence
  score so users can choose how strict to be.
  \item \textbf{The path to gold quality.} A human-correction layer on the same
  row identifiers upgrades every pair that uses a corrected column, and we
  propose how to test its effect on trained models.
\end{itemize}

The next section positions BharatSetu against existing multilingual MT corpora, models and benchmarks.

\section{Related Work}
\label{sec:related}
This section identifies the closest resource families and explains why BharatSetu fills a different slot: all 22 scheduled languages aligned to the same English rows in one table. Indic NLP resource building has progressed through monolingual corpora, benchmarks and multilingual models. IndicNLPSuite introduced monolingual corpora, evaluation benchmarks and pretrained multilingual language models for 11 Indian languages \citep{kakwani2020indicnlpsuite}. Samanantar collected large English--Indic parallel corpora for 11 Indic languages and used an English-centric corpus to derive Indic--Indic pairs through pivoting \citep{ramesh2022samanantar}. These resources established the importance of scale and shared infrastructure for Indic NLP, but they do not provide a single 22-language scheduled-language table with one row per English source sentence.

Recent model work has moved closer to full scheduled-language coverage. IndicTrans2 targets all 22 scheduled Indian languages and emphasises accessible MT models and data for those languages \citep{gala2023indictrans2}. Sarvam Translate is documented by Sarvam AI as a translation model supporting English--Indian-language translation across the official Indian-language set, and the BharatSetu generation process used the \texttt{sarvamai/sarvam-translate} model family \citep{sarvam2026translate}. BharatSetu is complementary to such systems: it is a derived data resource for training, evaluation design, auditing and correction workflows, not a new translation model.

Global multilingual MT research has also demonstrated the importance of many-language coverage and shared benchmarks. M2M-100 moved beyond English-centric translation by training a many-to-many multilingual model \citep{fan2021m2m100}. mBART showed that multilingual sequence-to-sequence pretraining can support MT, including low-resource settings \citep{liu2020mbart}. NLLB-200 and FLORES-200 scaled model and evaluation coverage to 200 languages and more than 40,000 translation directions, with human-translated evaluation data and safety analysis \citep{nllbteam2024scaling}. More recent Omnilingual MT work extends the ambition of multilingual MT beyond the 200-language scale \citep{omnilingual2026omt}. BharatSetu differs from these efforts in scope: it focuses on the administratively defined set of 22 scheduled Indian languages and on a uniform English-source row alignment rather than on a universal MT system.

Synthetic and machine-translated corpora have a long role in low-resource MT, especially when direct bitext is scarce. Back-translation and self-training are common ways to use model-generated text for training, but they require careful evaluation because machine-generated data can propagate errors and domain bias \citep{sennrich2016backtranslation,edunov2018backtranslation}. BharatSetu follows this tradition only as a machine layer. Its intended contribution is a structured starting point for experiments and correction, not a claim that synthetic translations are equivalent to human reference translations. Section~\ref{sec:construction} therefore describes the construction process before Section~\ref{sec:description} specifies the table users will inspect.

\section{Dataset Construction}
\label{sec:construction}
This section describes how BharatSetu preserves row alignment from English source rows through translation, merge and release packaging. The construction process starts from an English source corpus assembled from sources such as literature and news, alongside resources such as Wikipedia-derived WikiMatrix, OPUS, C4 and Tatoeba \citep{schwenk-etal-2021-wikimatrix,tiedemann2012opus,dodge-etal-2021-documenting,tiedemann-2020-tatoeba}. Each original row is represented by an \texttt{id}, an English sentence and a \texttt{source} provenance label. The \texttt{id} field is retained so that translations can be traced back to the original source-corpus row and so that future correction passes can update a row without changing its identity.

For each English row, translations were generated into the 22 scheduled-language target columns using the \texttt{sarvamai/sarvam-translate} model family. The resulting files were merged into a wide table with one row per English sentence. This wide format is important because it keeps all target-language outputs for a sentence adjacent, making cross-language comparison, correction and pair extraction a row-local operation.

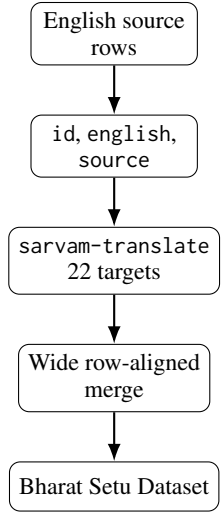
\begin{figure}[t]
\centering
\begin{tikzpicture}[
  node distance=0.65cm,
  box/.style={draw, rounded corners, align=center, minimum width=2.2cm, minimum height=0.65cm, font=\small},
  arrow/.style={-Latex, thick}
]
\node[box] (src) {English source\\rows};
\node[box, below=of src] (prov) {\texttt{id}, \texttt{english},\\\texttt{source}};
\node[box, below=of prov] (mt) {\texttt{sarvam-translate}\\22 targets};
\node[box, below=of mt] (merge) {Wide row-aligned\\merge};
\node[box, below=of merge] (pack) {Bharat Setu Dataset};
\draw[arrow] (src) -- (prov);
\draw[arrow] (prov) -- (mt);
\draw[arrow] (mt) -- (merge);
\draw[arrow] (merge) -- (pack);
\end{tikzpicture}
\caption{BharatSetu construction pipeline. The release keeps source-row identity, machine-generated target columns and packaging metadata separate so that provenance, translation quality and integrity can be audited independently.}
\label{fig:pipeline}
\end{figure}

Figure~\ref{fig:pipeline} summarises the construction pipeline. The figure intentionally distinguishes source provenance, machine translation and packaging, because these layers support different future checks.

The release contains sharded Parquet files under \texttt{data/train-*.parquet}, a single-file Parquet export, JSONL and XLSX copies, \texttt{metadata.json} and \texttt{generation/merge\_manifest.json}. These files serve different users: Parquet supports efficient training and filtering, JSONL supports streaming and line-based tooling, XLSX supports manual inspection, and metadata files support release review.

No human verification is claimed for the translations. This construction choice affects all downstream claims. BharatSetu can support experiments, sampling and correction workflows immediately, but it should not be used as a gold reference corpus without additional validation. The next section specifies the schema and language coverage that downstream users will see.

\section{Dataset Description}
\label{sec:description}
This section defines the table, columns and language coverage in BharatSetu 22-Indic 1M. Table~\ref{tab:schema} shows the schema. The required invariant is that each row contains one English source sentence and the 22 generated translations aligned to that sentence.

\begin{table}[t]
\centering
\small
\begin{tabular}{>{\raggedright\arraybackslash}p{0.24\linewidth}>{\raggedright\arraybackslash}p{0.63\linewidth}}
\toprule
\textbf{Column} & \textbf{Description} \\
\midrule
\texttt{id} & Identifier aligned to the original source-corpus row. \\
\texttt{english} & English source sentence. \\
\texttt{source} & Provenance label from the input corpus. \\
Target language columns & One machine-generated translation column per target language, named by the full language name. \\
\bottomrule
\end{tabular}
\caption{BharatSetu row schema. Each row contains source identity, English text, source provenance and 22 target-language columns.}
\label{tab:schema}
\end{table}

Table~\ref{tab:languages} lists the target languages and codes supplied in the fact sheet. Script conventions are not asserted as dataset facts because several scheduled languages have multiple orthographic traditions or release-specific script choices. Script handling should therefore be treated as a release-level audit item rather than an assumed property of the model output.

\begin{table*}[t]
\centering
\small
\begin{tabular}{lll|lll}
\toprule
\textbf{Language} & \textbf{Code} & \textbf{Layer} & \textbf{Language} & \textbf{Code} & \textbf{Layer} \\
\midrule
Assamese & as & machine & Bengali & bn & machine \\
Bodo & brx & machine & Dogri & doi & machine \\
Gujarati & gu & machine & Hindi & hi & machine \\
Kannada & kn & machine & Kashmiri & ks & machine \\
Konkani & kok & machine & Maithili & mai & machine \\
Malayalam & ml & machine & Manipuri & mni & machine \\
Marathi & mr & machine & Nepali & ne & machine \\
Odia & or & machine & Punjabi & pa & machine \\
Sanskrit & sa & machine & Santali & sat & machine \\
Sindhi & sd & machine & Tamil & ta & machine \\
Telugu & te & machine & Urdu & ur & machine \\
\bottomrule
\end{tabular}
\caption{Language coverage of BharatSetu. The 22 listed languages correspond to the Eighth Schedule set. Script conventions in the generated columns should be audited by the authors before public release.}
\label{tab:languages}
\end{table*}

The corpus contains 984,764 English source rows and 22 target-language columns, yielding 21,664,808 machine-generated Indic cells. The public dataset is hosted on \href{https://huggingface.co/datasets/Omarrran/BharatSetu-22Indic-1M}{Hugging Face}~\citep{bharatsetu2026} and distributed in sharded Parquet, single-file Parquet, JSONL and XLSX formats.

The row count and target count imply the pair structure used later. There are 22 English-to-Indic directions and 22 Indic-to-English directions if the English column is used as a target, for 44 English-connected directions. For Indic-to-Indic use, each source language can be paired with each of the other 21 target languages, giving $22\times21=462$ directed pivot-derived pairs. The total directed directions available from the aligned table are therefore $44+462=506$. Section~\ref{sec:uses} explains why this arithmetic matters for actual research workflows rather than only for dataset description.

\section{Why Row-Aligned Multiway Data Solves Real Problems}
\label{sec:uses}
Row alignment makes one table useful for training, pivoting, transfer, analysis and distillation across all scheduled languages. A bilingual corpus supports one source--target direction; a collection of unrelated bilingual corpora supports several, but not on shared domains or sentences. A row-aligned table supports controlled extraction of many directions from one sentence spine.

The first use case is English-to-22 and 22-to-English training. Every row has an English source and one target column per scheduled language, so a user can build 22 English-to-Indic training files with identical row identifiers and provenance labels, and reverse each for Indic-to-English training. The Indic side is machine-generated, whereas the English side is original source text. Using the same rows for every language reduces confounding by row selection when languages are compared.

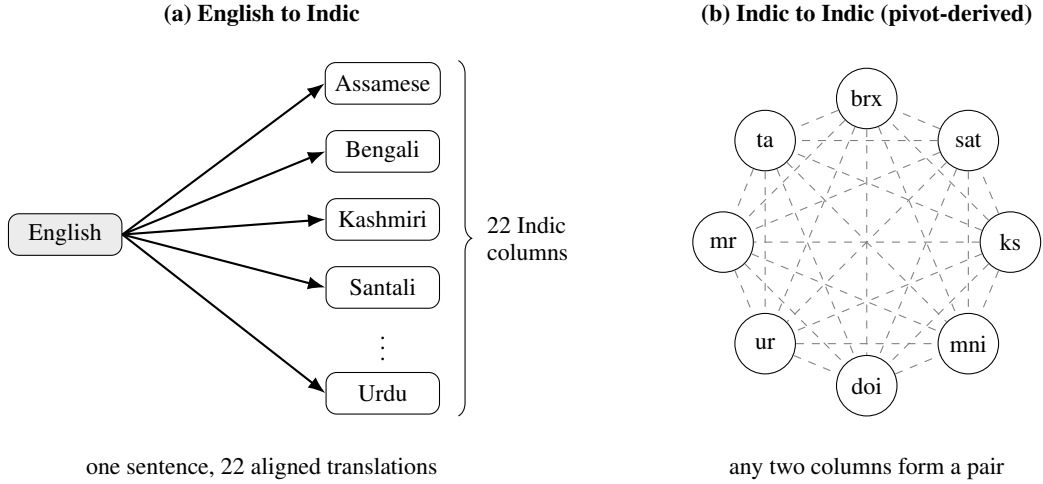
\begin{figure*}[t]
\centering
\begin{tikzpicture}[
  font=\small,
  cell/.style={draw, rounded corners, align=center, minimum width=1.5cm, minimum height=0.55cm},
  lang/.style={draw, circle, inner sep=0pt, minimum size=0.8cm, font=\footnotesize, fill=white},
  src/.style={-Latex, thick},
  pair/.style={dashed, thin, gray}
]
\node[font=\small\bfseries] at (2.6,3.3) {(a) English to Indic};
\node[cell, fill=gray!15] (en) at (0,0.4) {English};
\node[cell] (as)  at (4.2,2.4)  {Assamese};
\node[cell] (bn)  at (4.2,1.5)  {Bengali};
\node[cell] (ks)  at (4.2,0.6)  {Kashmiri};
\node[cell] (sat) at (4.2,-0.3) {Santali};
\node        (dt)  at (4.2,-1.0) {$\vdots$};
\node[cell] (ur)  at (4.2,-1.7) {Urdu};
\foreach \l in {as,bn,ks,sat,ur} \draw[src] (en.east) -- (\l.west);
\draw[decorate, decoration={brace, amplitude=5pt}] (5.2,2.7) -- (5.2,-2.0)
  node[midway, right=7pt, align=left, font=\footnotesize] {22 Indic\\columns};
\node[font=\footnotesize, align=center] at (2.6,-2.7) {one sentence, 22 aligned translations};

\node[font=\small\bfseries] at (10.6,3.3) {(b) Indic to Indic (pivot-derived)};
\foreach \i/\n in {0/ks, 1/sat, 2/brx, 3/ta, 4/mr, 5/ur, 6/doi, 7/mni}
  \node[lang] (n\i) at ({10.6+1.9*cos(45*\i)},{0.3+1.9*sin(45*\i)}) {\n};
\foreach \i in {0,...,7}{\foreach \j in {0,...,7}{\ifnum\i<\j \draw[pair] (n\i) -- (n\j);\fi}}
\node[font=\footnotesize, align=center] at (10.6,-2.7) {any two columns form a pair};
\end{tikzpicture}
\caption{The two stages of the row-aligned design. (a)~English to 22 Indic columns. (b)~Pivot-derived Indic-to-Indic pairs. Eight of the 22 languages are shown in~(b) by ISO code; dashed lines apply in both directions.}
\label{fig:en2indic2indic}
\end{figure*}

The second use case is Indic-to-Indic translation through an English pivot (Figure~\ref{fig:en2indic2indic}). Panel~(a) shows the construction: each English sentence is translated into the 22 Indic targets, giving one English cell and 22 aligned Indic cells per row. Panel~(b) shows the consequence: any two Indic cells in a row can be paired, for example Kashmiri--Santali or Bodo--Tamil, with no further translation step. These are \emph{pivot-derived pairs}, not independently human-translated bitext. Their semantic link is mediated by the shared English source, and because both sides are machine translations, errors on either side can carry into the pair; pair quality is therefore a question to be evaluated. For 22 Indic targets there are $22\times21=462$ directed Indic-to-Indic pairs; adding the 44 English directions (22 in each direction) gives 506, which equals $23\times22$ for 23 aligned columns.

The third use case is transfer for lower-resource scheduled languages. Bodo, Dogri, Santali, Kashmiri and Manipuri are often harder to serve with large parallel corpora than Hindi, Bengali, Tamil or Telugu. A uniform table lets researchers test whether multilingual training transfers representations from higher-resource to lower-resource columns under a shared row distribution; this is an open empirical question, not a result of the corpus.

The fourth use case is controlled cross-lingual analysis. Because every target cell in a row aligns to one English sentence, researchers can compare how languages handle named entities, numerals, punctuation, honorifics, morphology and word order. The same structure supports language-identification experiments, transliteration studies, script-conversion audits and consistency checks for multilingual generation, with the assurance that each row shares one source meaning even when target quality varies.

The fifth use case is synthetic-data distillation. A model trained on BharatSetu can be compared with the original translation model family, an IndicTrans2 baseline, or a general multilingual MT model such as NLLB-200, to ask whether a compact or domain-adapted model recovers useful performance from a uniform synthetic table, and whether it inherits or amplifies errors from the machine layer. This leads to the human correction roadmap in Section~\ref{sec:correction}.

\section{Human Correction Layer: Roadmap}
\label{sec:correction}
This section proposes a correction design that improves the machine layer without misrepresenting the current release as human-verified. The simplest model is layered. Tier 0 is the released machine layer: 984,764 rows, 22 target columns and no human verification claim. Tier 1 is a sampled human-verified evaluation layer for selected languages and domains. Tier 2 is a set of fully corrected priority-language columns or row subsets. Tier 3 is a versioned release in which corrected cells coexist with machine cells and preserve the same row identifiers.

The key advantage of a correction layer is that each corrected cell has network effects. If the Tamil column is corrected for a row, then English--Tamil improves directly and every pivot-derived pair involving Tamil receives a better Tamil side. If a new corrected Kashmiri subset is added, it improves English--Kashmiri, Kashmiri--English and all Kashmiri--other-language pivot-derived pairs for those rows. The benefit scales by connection rather than by one isolated bilingual file.

Human correction should be implemented as an auditable annotation process. The authors should define language-specific guidelines, annotator qualifications, compensation policy, adjudication rules, version identifiers and inter-annotator agreement measures. For languages with script variation, the guidelines must specify the expected script or permit multiple scripts with explicit labels. Corrected columns should not overwrite the machine columns silently; they should be stored as additional versioned fields or release variants so that experiments can compare the machine layer against the human correction layer.

The main research hypotheses are as follows. First, training on the machine layer may outperform having no in-domain or no aligned data for lower-resource scheduled languages. Second, training on a corrected subset may improve quality relative to training on the machine layer alone. Third, correcting a small number of strategically chosen languages may improve pivot-derived pairs that use those languages. These hypotheses require measurement before they can be asserted.

\section{Recommended Evaluation Protocol for Users}
\label{sec:evaluation}
This section reports no results. It gives a protocol that we recommend to anyone who uses BharatSetu, whether for training, evaluation or analysis, because the corpus is machine-generated and its quality has not been measured. The protocol combines data-integrity checks, automatic MT metrics, human evaluation and correction-ablation experiments. Where reference translations exist, users should score with BLEU, chrF and COMET \citep{papineni2002bleu,popovic2015chrf,rei2020comet}, and report metric choices with caveats for morphology, script variation and domain mismatch. Human evaluation should cover adequacy, fluency, error typology and harmful-content or bias checks, following error-analysis practice such as MQM where qualified annotators are available \citep{freitag2021experts}.

Table~\ref{tab:evaluation} summarises the recommended protocol. We advise running every experiment on a fixed dataset version and documented model checkpoints.

\begin{table*}[t]
\centering
\small
\setlength{\tabcolsep}{3pt}
\begin{tabular}{>{\raggedright\arraybackslash}p{0.17\linewidth}>{\raggedright\arraybackslash}p{0.21\linewidth}>{\raggedright\arraybackslash}p{0.26\linewidth}>{\raggedright\arraybackslash}p{0.27\linewidth}}
\toprule
\textbf{Evaluation target} & \textbf{Directions} & \textbf{Data condition} & \textbf{Metrics or judgments} \\
\midrule
Machine-layer MT quality & English$\rightarrow$22 and 22$\rightarrow$English & Released machine layer; fixed sample per language & chrF, BLEU, COMET where references exist; human adequacy and fluency \\
Pivot-derived pairs & Selected low-resource and high-resource non-English pairs & Pairs derived from shared English rows & Human adequacy; error-propagation categories \\
Confidence layer & Same non-English pairs & Tier 0, Tier 1 and Tier 2 subsets of one pair & Metric deltas on a fixed human-checked test set \\
Correction benefit & Priority languages and sampled rows & Machine layer vs.\ human-corrected subset & Metric deltas and paired human preference \\
Robustness & Domain, length and script buckets & Stratified samples using \texttt{source}, length and script labels & Error rates by bucket; qualitative audit \\
Responsible-use audit & All languages, sampled & Released and corrected variants & Bias, toxicity, sensitive named-entity and dialect checks \\
\bottomrule
\end{tabular}
\caption{Recommended evaluation protocol for users of BharatSetu. No results are reported in this paper.}
\label{tab:evaluation}
\end{table*}

Users who train models should run four comparisons. The first compares models trained with no BharatSetu data against models trained on the released machine layer. The second compares machine-layer training against training on a human-corrected subset. The third compares Tier~0, Tier~1 and Tier~2 subsets of the same language pair (Section~\ref{sec:correction}). The fourth tests whether correcting one language column improves the pivot-derived pairs that use it. Every experiment should record the repository revision, source rows, language columns, preprocessing, tokenizer, model checkpoints, metric versions and random seeds. This record matters because the corpus is synthetic and because a small change in filtering or correction can change how results are interpreted.

Before using the data, we also recommend dataset-integrity checks: each release file should have 984,764 rows and the expected 25 columns, no row should be shuffled across target columns, and shard counts should match \texttt{generation/merge\_manifest.json}. These checks establish whether the resource is structurally safe to use, not translation quality; Section~\ref{sec:limitations} states what users still cannot infer once they pass.

\section{Limitations and Ethics}
\label{sec:limitations}
This section states the limits that constrain all claims about BharatSetu. The corpus is machine-translated and not human-verified. Translation errors may include mistranslation, omission, hallucination, named-entity corruption, register mismatch, grammatical errors and script inconsistency. Quality may vary by language, domain and sentence style. These limitations are especially important for lower-resource languages, where model performance may be less stable and where users may have fewer independent references for checking outputs.

The row-aligned design creates pivot-derived pairs, but it does not create direct human-translated bitext between non-English languages. A Tamil--Kashmiri pair derived from a shared English row may be useful for training or analysis, yet it can inherit errors from both generated columns and from English-centric semantic framing. Researchers should label such data as pivot-derived and should avoid using it as a gold reference for evaluation unless a human correction layer has been added.

The resource may also inherit social bias, dialect imbalance and domain skew from both the English source corpus and the translation model. Scheduled-language labels do not fully capture dialect diversity, register variation or script choice within each language. Responsible use therefore requires evaluation with native speakers, transparent error reporting and care in high-stakes settings such as legal, medical, educational or public-service translation.

\section{Conclusion}
India's 22 scheduled languages have uneven parallel data, and most language pairs among them have no direct bitext. BharatSetu addresses this by placing 984,764 English sentences and their machine translations into all 22 languages in one row-aligned table. Because every Indic cell shares an English source, any two columns form a pivot-derived pair, which connects any Indic language to any other without direct data. A confidence layer lets users choose how strictly each pair is connected. The corpus is machine-generated and not human-verified, and pivot-derived pairs can inherit errors from both sides, so we report no quality results. We release BharatSetu as a machine layer and propose a versioned human-correction layer and an evaluation protocol. Each corrected column would upgrade every pair that uses it, so the table can improve step by step into stronger evidence for multilingual Indic machine translation. Future work is to run the protocol, starting with the lower-resource languages.

\bibliography{bharatsetu_22indic_1m}

\end{document}